\documentclass[runningheads]{llncs}
\usepackage{graphicx}
\usepackage{import}
\usepackage[nogroupskip=true, style=super]{glossaries}

\newacronym{nlp}{NLP}{Natural Language Processing}
\newacronym{vqa}{VQA}{Visual Question Answering}
\newacronym{vl}{VL}{Visual-Linguistic}
\newacronym{lpf}{LPF}{Linguistic Processing Flow}
\newacronym{vpf}{VPF}{Visual Processing Flow}
\newacronym{fpt}{FPT}{Fusion and Pre-Training}
\newacronym{ml}{ML}{Machine Learning}
\newacronym{1g}{1G}{First Generation}
\newacronym{2g}{2G}{Second Generation}
\newacronym{vcr}{VCR}{Visual Commonsense Reasoning}
\newacronym{dl}{DL}{Deep Learning}
\newacronym{lstm}{LSTM}{Long-short Term Memory}
\newacronym{cnn}{CNN}{Convolutional Neural Network}
\newacronym{ffn}{FFN}{Feed Forward Network}
\newacronym{mlm}{MLM}{Masked language Model}
\makeglossaries

\usepackage{soul,color}
\usepackage{xcolor}
\usepackage{ulem}

\begin{document}
%
%\title{Contribution Title\thanks{Supported by organization x.}}
\title{A Multimodal Memes Classification: A Survey and Open Research Issues}
	
\author{Tariq Habib Afridi \inst{1}\orcidID{0000-0002-1535-8203} \and
Aftab Alam\inst{1}\orcidID{0000-0001-9222-2468} \and
Muhammad Numan Khan\inst{1}\orcidID{0000-0001-5892-9064} \and
Jawad Khan \inst{1}\orcidID{0000-0001-8263-7213} \and
Young-Koo~Lee\inst{1}}
%Young-Koo~Lee\inst{1}\orcidID{2222--3333-4444-5555}}
%
\authorrunning{T.H. Afridi et al.}
\institute{Department of Computer Science and Engineering, Kyung Hee University (Global Campus), Yongin 1732, South Korea \\
\email{\{afridi,aftab,numan,jkhanbk1,yklee\}@khu.ac.kr}}
%\and
%School of Computer Science and Communication Engineering, Jiangsu University \\
%\email{\{shahkhalid\}@ujs.edu.cn}}
%
\maketitle              % typeset the header of the contribution
\begin{abstract}
Memes are graphics and text overlapped so that together they present concepts that become dubious if one of them is absent. It is spread mostly on social media platforms, in the form of jokes, sarcasm, motivating, etc. After the success of BERT in \gls{nlp}, researchers inclined to \gls{vl} multimodal problems like memes classification, image captioning, \gls{vqa}, and many more.  Unfortunately, many memes get uploaded each day on social media platforms that need automatic censoring to curb misinformation and hate. Recently, this issue has attracted the attention of researchers and practitioners. State-of-the-art methods that performed significantly on other \gls{vl} dataset, tends to fail on memes classification. In this context, this work aims to conduct a comprehensive study on memes classification, generally on the \gls{vl} multimodal problems and cutting edge solutions. We propose a generalized framework for \gls{vl} problems. We cover the early and next-generation works on \gls{vl} problems. Finally, we identify and articulate several open research issues and challenges.  This is the first study that presents the generalized view of the advanced classification techniques concerning memes classification to the best of our knowledge. We believe this study presents a clear road-map for the \gls{ml} research community to implement and enhance memes classification techniques.

%The abstract should briefly summarize the contents of the paper in 15--250 words.

\end{abstract}

\section{Introduction}
\label{sec:Introduction}
%Motivation and Background
From the past couple of years, there has been an elevation in the research community on multimodal problems such as \gls{vqa} \cite{antol2015vqa,goyal2017making,wang2020visual} and image captioning \cite{chen2015microsoft,gurari2020captioning,wang2020visual}, memes classification \cite{kiela2020hateful}, and many more \cite{mogadala2019trends}. Many real-world problems are multimodal, just like humans perceive the world using multimodal senses such as eyes, ears, and tongues. Likewise, data on the internet and machine are also multimodal, which is in text, image, video, sound, etc. Memes classification is a multimodal problem as most of the memes have two modalities, such as text and graphics from the image. Due to the heavy use of social media platforms, there has been a requirement to curb the negative impact automatically. One such problem is the automatic filtering of hateful memes to stop the users from spreading hate across the internet. Big social media platforms like Twitter, Facebook are often instructed by different countries to stall the spread of online hatred. Facebook has recently called for memes classification challenges2020 \cite{kiela2020hateful}, which includes hateful memes such as racist, sexist, and some may incite violence.

%Problems Statement
Recent \gls{vl} multimodal techniques such as \cite{li2019visualbert,lu2019vilbert} have been found far from human accuracy and multimodal memes classification is still in its infancy \cite{kiela2020hateful}. This is a challenging problem because separately, a meme can have a pleasant caption and a normal picture, but it may become offensive when combined in a certain way. Consider a meme that has a caption like “love the way you smell today". Combine that caption with the image of a skunk, and it became mean. Similarly, consider a caption "look how many people love you" which seems good but adds that with barren land, it also becomes mean. Likewise, consider an image of women face beaten and add that with the texts such as "women ask for equal rights, so I give them equal lefts aswell" and then "women ask for equal rights, and this is why". Now the challenge will for the vision model be to identify beaten women to classify these two memes as hateful/non-hateful. This brings up the need for multimodal models for training jointly on the text and images simultaneously. To classify memes accurately, we will need the \gls{vl} multimodal models to understand the concepts in memes which otherwise need human intelligence. It will also need a greater intelligent model for general \gls{vl} multimodal problems \cite{mogadala2019trends}.

%State of the art trends
Since the vision \cite{krizhevsky2012imagenet,simonyan2014very,szegedy2015going,he2016deep}, and language \cite{devlin2018bert,liu2019roberta,lan2019albert} tasks independently has seen a lot of progress in recent years. Unfortunately, they are still lagging on multimodal problems like \gls{vqa}, image captioning, hateful memes classification, etc. In the next few years, we may see a peak on \gls{vl} multimodal problems for the research community. Further, recent research on multimodal has been found with similar kinds of problems such as language can carelessly enforce strong priors that can end in an outwardly impressive performance, neglecting the core model's visual content \cite{devlin2015exploring}. Related issues can be found in \gls{vqa} \cite{antol2015vqa}, where without refined multimodal, a simple baselines model performed unusually well \cite{zhou2015simple,agrawal2016analyzing,goyal2017making} that will unlikely work on memes classifications as well.

%Contributions, Aim, and Objectives
This work aims to conduct a thorough investigation on the status of advanced \gls{ml} approaches on memes classification in social media. First, a generic framework has been proposed for social media memes classification. Then we present a broad overview of up-to-date, relevant literature. Eventually, open research issues and challenges are addressed, with a focus on the proposed framework.

The remaining paper is organized in the following order. Section~\ref{sec:GenericFramework} discusses a generic memes classification framework. In section~\ref{sec:ResearchLiterature}, a recent literature review has been discussed. Section~\ref{sec:ResearchIssues} presents several open research issues and Challenges. Finally, we conclude this study in Section~\ref{sec:Conclusion}.

\section{Memes Classification: A generic architecture}
\label{sec:GenericFramework}
\begin{figure*}[htbp]
    \centerline{\includegraphics[width=\textwidth]{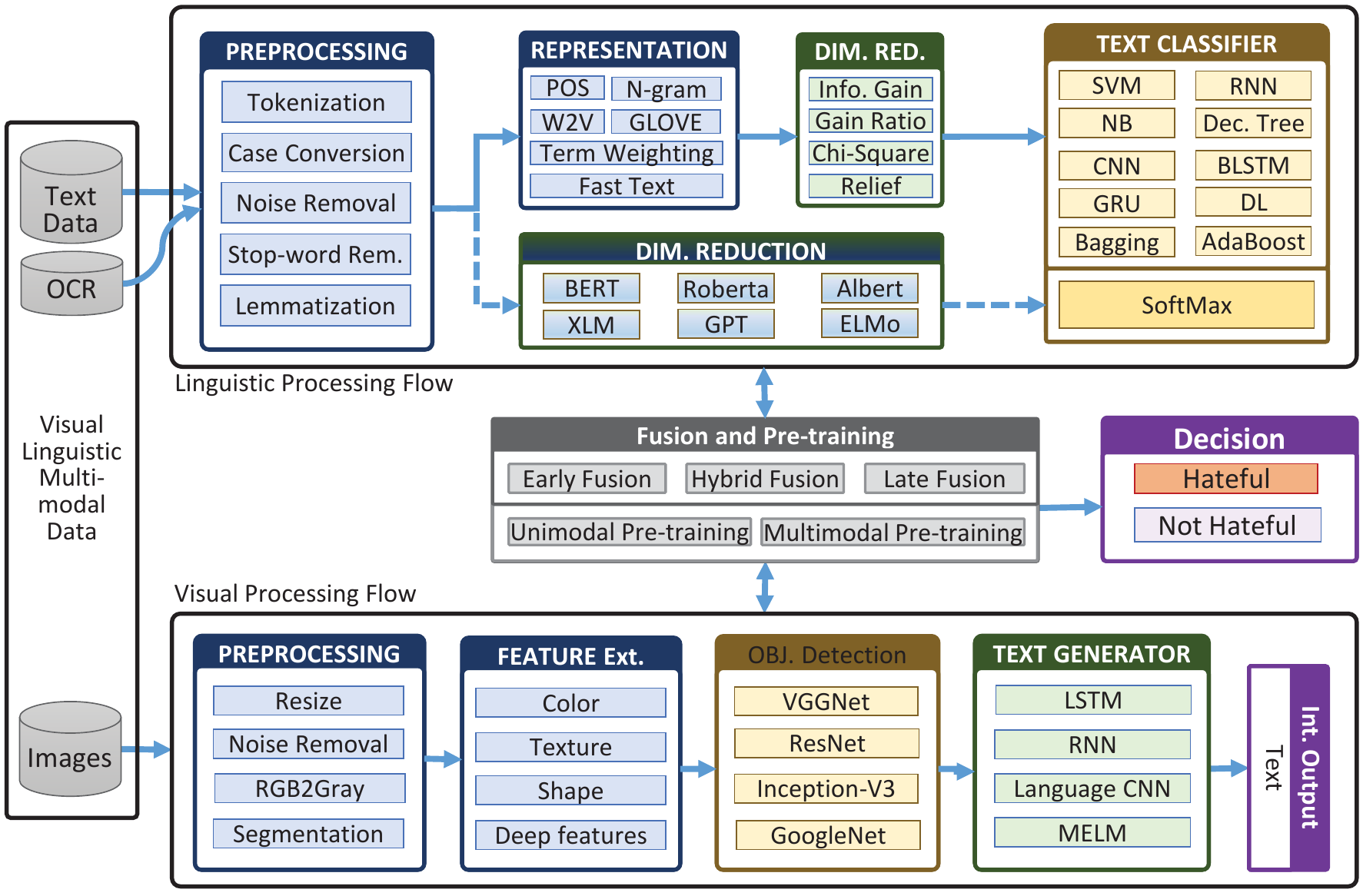}}
    \caption{A generic architecture for memes classification.}
    \label{fig:ArchitectureMemes}
\end{figure*}

The memes classification task can be seen as a combined \gls{vl} multimodal problem. It is different from some of the current \gls{vl} problems like image captioning, where efforts are made to find the best possible explanations for the image in the form of the caption, whereas, in memes, we have to make the decision based on semantically correlated text with that of the visual content in the image. Therefore, a cross-modal approach under vision and text will only perform better on memes classification. Traditional \gls{vl} approaches were based on simple fusion in the form of early or late fusion while unimodally learning each vision and language problems. However, a multimodally pre-trained model may perform better to classify memes. Based on an extensive literature review, we proposed a generic multimodal architecture for memes classification shown in Figure~\ref{fig:ArchitectureMemes}. The proposed architecture has two types of flows, i.e., \gls{lpf} and \gls{vpf}.  There is a middle phase called \gls{fpt}, which will define the fusion and Pre-training strategies for merging \gls{lpf}, and \gls{vpf} \cite{baltruvsaitis2018multimodal,kiela2020hateful}. 

Both \gls{nlp} and vision has a long history of \gls{ml} methods, for which we have categorized them in the first and second generation. After the success of AlexNet \cite{krizhevsky2012imagenet}, the next generation of vision has begun, which is based solely on deep learning models, specifically convolutional neural networks. Similarly, After the success of Bert \cite{devlin2018bert}, the next generation of \gls{nlp} also gets started. We divided both \gls{lpf} and \gls{vpf} flow into two generations; that is \gls{1g} and \gls{2g}. We will further elaborate on each of them in the following subsections.

\subsection{Linguistic Processing Flow}
In \gls{1g} \gls{lpf}, embedding's can mostly capture the semantic meanings of words. However, such techniques are context unaware and fail to capture higher-level contextual concepts, such as polysemous disambiguation, syntactic structures, semantic roles, and anaphora. Formally \gls{lpf}, considering \gls{1g}, is a four-step process. i.e., pre-processing, feature engineering, dimensionality reduction, and classification. Tasks like stop word removal, capitalization, tokenization, abbreviation handling, slang and idioms handling, spelling correction, noise removal, lemmatization, and stemming are performed in pre-processing \cite{kowsari2019text}. Fortunately, we will not need many of these sub-tasks as text on memes is observed mostly to be clean. However, non-English memes may need some other kind of pre-processing for cleaning linguistic defects. After pre-processing, feature engineering steps are performed to extract useful features from the text. Feature engineering is a non-trivial task as they have to look for better representation of the extracted features. Some popular feature engineering techniques being used are word embeddings like Word2Vec, GloVe, syntactic word representation like N-Gram, weighted words such as Bag of words (BoW), Term Frequency-Inverse Document Frequency (TF-IDF), and FastText \cite{kowsari2019text}. The noteworthy flaws in these techniques are that they failed to capture the context and correct meaning of words such as words with multiple meanings in a different context. 

Dimensionality reduction techniques like PCA, ICA, LDA, etc., are also employed by research communities to terminate unwanted features. Once the quality features are extracted, the most critical phase in \gls{1g} text classification pipeline is picking the best classification model. To determine the most effective classification model for any \gls{nlp} task, a conceptual understanding of each of these algorithms was a necessity. Therefore, researchers have employed typical text classifiers, such as SVM, kNN, Naïve Bayes, ensemble classifiers such as Bagging, Adaboost, and Decision Tree, Random Forest, which are tree-based \cite{kowsari2019text}. Recently, \gls{dl} methods have attained superior results compared to the earlier \gls{ml} algorithms on tasks such as object detection in image, face recognition, \gls{nlp}, etc. The reason for the success of these approaches depends on the ability to model the complex and non-linear associations inside the data.

%%%%%%%%%%%%%%%%%%%%%%%%%%%%%%%%%%%%%%%%%%%%%%%%%%%%%%%%%%%
%Second Generations?
In \gls{2g}, the focus of the research community in \gls{nlp} has been shifted to neural network-based approaches such as RNN, CNN, and transformer-based attention models such as BERT \cite{devlin2018bert}, OpenAI GPT-2 \cite{radford2019language}, Roberta \cite{liu2019roberta}, and Albert \cite{lan2019albert}. Since the inception of BERT, a new era has been started in \gls{nlp} as it attained state-of-the-art results on many \gls{nlp} tasks. BERT is built on top of several past clear ideas, and it incorporates ideas from semi-supervised sequence learning \cite{dai2015semi},  ElMo \cite{peters2018deep} (that solved the problem of Polysemy by using layers of complex Bi-directional \gls{lstm} architecture), ULMFiT \cite{howard2018universal} (which trained language models that could be fine-tuned to provide excellent results even with fewer data thus cracking the code for transfer learning in NLP), and substituting \gls{lstm} by the transformer \cite{vaswani2017attention} (gives better parallel processing and shorter training time than that of \gls{lstm}). The transformer from Vaswani \cite{vaswani2017attention} has enhanced the \gls{nlp} by capturing relationships and the sequence of words in sentences, which is vital for a machine to understand a natural language understanding. Unlike, \gls{1g} approaches, which heavily relied on feature engineering and choosing the best classifier, were a burdensome task. However, BERT has made the job easy as it is pre-trained on a huge corpus of data, and by consuming the transfer learning, it can be fine-tuned to any given task.

Attention layers \cite{bahdanau2014neural} from the transformer tends to align and extract information from a query vector using context vectors. Attention normalizes the calculated matching score between the query vector and each context vector among all vectors using softmax. Self-attention is an attention layer in which the input query vector is in the set of context vectors i.e it just replace the target sequence with the same input sequence. Explicitly, most researchers now tends to the multi-head attention \cite{vaswani2017attention}. The common transformer architecture is composed of encoders and decoders, which are a heap of several identical layers comprising of position-wise \gls{ffn} layer and multi-head self-attention layer. The temporal aspect of sequential input has been explored by the position-wise \gls{ffn} and is accounted for by the transformer in the encoder phase by generating content embedding and position encoding for each token of the input sequence. While, the self-attention within each sub-layer in the Multi-head Self-Attention is used to align tokens and their positions among the same input sequence. Sequence models usually capture the local context of a word in sequential order such as \gls{lstm}, which is common in language processing and generation among researchers. However, transformer architecture attains substantial parallel processing, reduced training time, and sophisticated accuracy for translation without any recurrent component, unlike \gls{lstm}, which is a remarkable advantage. On the contrary, weakly incorporated position information from the position encoding may perform worse for problems that are sensitive to positional variation.

\subsection{Visual Processing Flow}
After the success of AlexNet \cite{krizhevsky2012imagenet}, the focus has been shifted from traditional \gls{1g} that consists of old-fashioned steps such as pre-processing, feature engineering, and classification. In \gls{1g}, the researchers have an inept job exploring and redesigning the feature engineering process for any particular but slightly different problem or domain. They also had an additional load of choosing the best classification model for their generated features. Various feature extraction methods have been employed on images like LBP, SIFT, HOG, SURF, BRIEF, and many more \cite{kortlicomparative}. Similarly, many traditional classification methods have also been employed with zero transfer learning capability. Moreover, the traditional visual processing cycle required a similar work for finding the best methods in each sub-step, like pre-processing, feature extraction, feature selection, and classification. This issue has been overcome by CNN, to acquire features at its own and, at the same time, it can be fine-tuned to other related tasks by transfer learning.

\begin{figure*}[htbp]
    \centerline{\includegraphics[width=\textwidth]{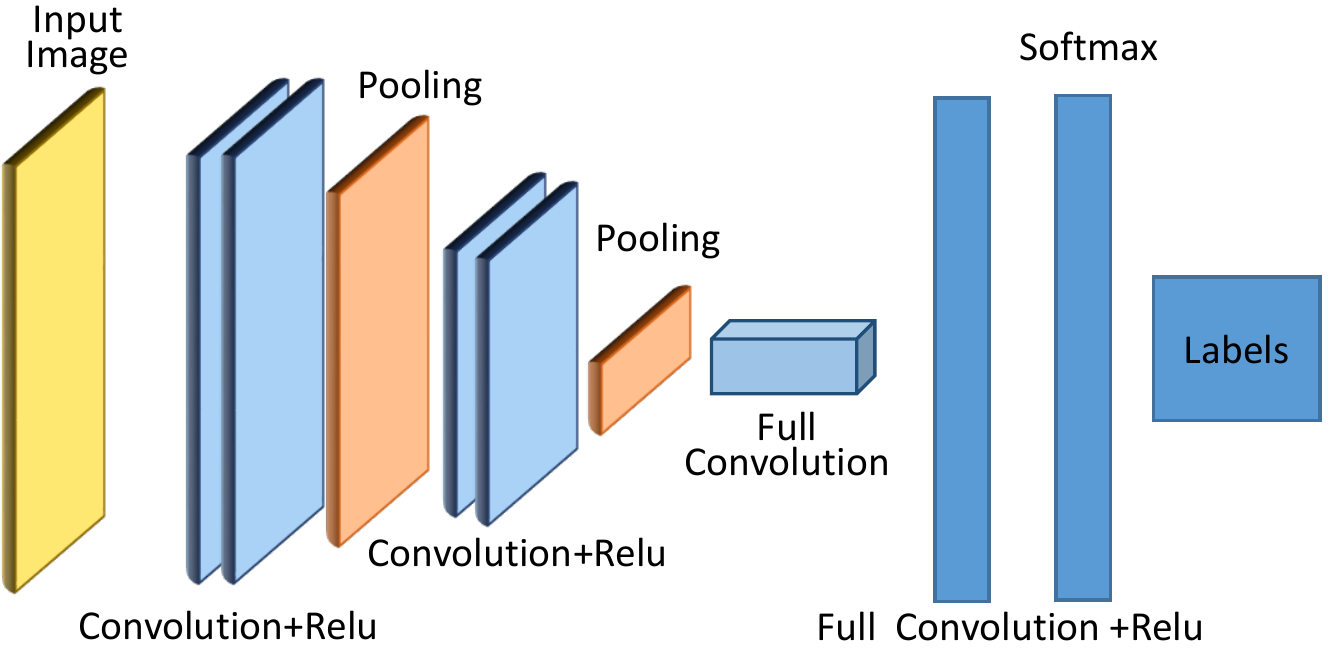}}
    \caption{A generic CNN Architecture.}
    \label{fig:CNNArchitecture}
\end{figure*}

ImageNet database \cite{deng2009imagenet} has changed the course of \gls{ml} by shifting the focus to deep learning altogether, and thus we called it the \gls{2g} era of \gls{vpf}. In \gls{dl}-based models, it can incorporate a large and diverse set of images and videos as they automatically learn features from training data and can be generalized well on related problems. The general architecture of \gls{cnn} can be seen in Figure~\ref{fig:CNNArchitecture}, they are mostly a combination of convolution layers plus activation functions like Relu and its variants, subsampling layers like max-pooling, fully convolutional layer, dense layer, and a softmax layer in the end. Since AlexNet \cite{krizhevsky2012imagenet} astonishing results in ImageNet challenge, computer vision research has been shifted to enhance the \gls{cnn} architecture. VGG \cite{simonyan2014very} and the inception module of GoogLeNet \cite{szegedy2015going} shown the benefits of expanding the depth and width of the \gls{cnn} architecture. ResNets \cite{he2016deep} developed the residual learning block by going through the shortcut connection of identity mapping, enabling the neural network model to burst through the obstruction of hundreds or even thousands of layers. DenseNet reformulates the connections between network layers that further boost the learning and representational properties of deep networks \cite{huang2017densely}. Moreover, extensive research has been done in object detection, and many \gls{cnn}-based approaches have been proposed, such as RCNN, Faster-RCNN, Yolo, etc. \cite{liu2020deep}.  

\subsection{Fusion and Pre-Training: towards Multimodality}
Based on the literature \cite{baltruvsaitis2018multimodal}, we categorize the \gls{vl} multimodal fusion into three categories, they are early fusion, late fusion, and hybrid fusion. Early fusion merges the features instantly after they are extracted. On the other end, late fusion integrates the decisions after each modality has taken its decision. Lastly, hybrid fusion fuse outputs from individual unimodal predictors and early fusion.
The pre-training consists of unimodally pre-trained and multimodally pre-trained. A unimodally pre-trained language and vision model combined by different fusion types is called a unimodally pre-trained multimodal. In contrast, the multimodally pre-trained language and vision model is called multimodally pre-trained \cite{kiela2020hateful}.

\section{State-of-the-art on Memes Classification}
\label{sec:ResearchLiterature}
Since there isn't much work done on multimodal memes classification, we consider other \gls{vl} problems for state-of-the-art. A classic task in the \gls{vl} multimodal exploration is to understand an alignment between multimodal feature spaces. Generally, in this context, a \gls{cnn} and an RNN are trained together to learn a combined embedding space from aligned multimodal \gls{vl} data and is a commonly followed architecture in image captioning \cite{hossain2019comprehensive,gurari2020captioning}. Contrastively, \gls{vqa} merges both \gls{vl} modalities to decide the right answer instead of learning an alignment between two spaces. It requires the precise correlations modeling between the image and the question representations. In hateful memes, a similar kind of accurate correlation modeling between image and texts is required as we need to find suitable alignment to both \gls{vl} modalities to comprehend the underlying correlation among modalities and finally make a decision. For memes, we take inspiration from the \gls{vqa} literature for the state-of-the-art models \cite{srivastava2019visual}.

At the beginning of the \gls{vqa}, researchers employed early fusion by feature concatenation. Later methods learned multimodal features using bilinear pooling \cite{fukui2016multimodal}. These methods have severe limitations as the multimodal features are fused in the latter stage of the model, so the alignment of \gls{vl} was also weakly extracted. Also, the acquired visual features by demonstrating the output of \gls{cnn} as a one-dimensional vector significantly losses the spatial information from the input image \cite{gomez2020exploring}. Recently, the focus has been shifted to cross-modality by multimodal pre-training approaches like Visualbert \cite{li2019visualbert}, UNITER \cite{chen2019uniter}, and Vilbert \cite{lu2019vilbert}. They have outclassed many recent approaches on multiple \gls{vl} multimodal datasets such as \gls{vqa} \cite{antol2015vqa}, \gls{vcr} \cite{zellers2019recognition}, NLVR \cite{suhr2018corpus}, Flicker30K \cite{plummer2015flickr30k}, and many more.

% Text Hateful NLP work
\subsection{Hateful speech Classification}
Considerable work has been carried out in recent years on detecting the hate speech \cite{fortuna2018survey}. Many techniques have been proposed by researchers varying in the feature engineering domains as well as in the choosing of classifiers. Traditional feature engineering techniques like BOW, N-grams, POS, TF-IDF, CBOW, word2vec, and text features have been employed for hate speech detection. Similarly, various classification algorithms have also been employed, out of which most frequents are SVM, Random Forest, Decision tree, Logistic regression, Naïve Bayes, and many more \cite{fortuna2018survey}. Unlike other tasks in \gls{nlp}, hate speech has a taste of cultural and regional implications; subjected to one specific cultural background, any expression may be professed as offensive or not. Also, hate speech detection in English by well-known methods can be seen as how correspondingly effective they are in other languages \cite{schmidt2017survey}.

Very little \gls{2g} \gls{nlp} methods have been employed on hate speech from social media sites \cite{mozafari2019bert,sohn2019mc}. One such method has used BERT by using a new fine-tuning approach based on transfer learning to capture hateful content within social media posts \cite{mozafari2019bert}. These fine-tuning were; initially with minimum changes, then inserting nonlinear layers, and finally, with inserting the Bi-\gls{lstm} layer and \gls{cnn} layer. They achieved the best result in the insertion of \gls{cnn} layers for fine-tuning. Another method proposed a multi-channel BERT model employing three BERT, one for the multilingual task, one for the English, and one for the Chinese \cite{sohn2019mc}. They explored the translations capabilities by interpreting training and test sentences to the equivalent languages requisite by these three different BERT models. They also evaluated their model on three non-English and non-Chinese language datasets and compared their previous methods approach. Further, they used Google Translation API to translate the text of the source language to English and Chinese for feeding that into corresponding English and Chinese BERTs. Lastly, they compared and presented the state of the art performance using their model.
% Multimodal V-L Classification
\subsection{Multimodal Visual-Linguistic Classification}
Substantial research has been conducted in the past decade on integrating the vision and language modalities. Most of these \gls{vl} models have similar architectures as they are generally pre-trained \gls{cnn} models for a variety of computer vision tasks ranging from scene recognition to object detection and object relation among them as well. Likewise, for the language representation, most of these models employed RNN, specifically LSTM and GRU being the most popular choices in the near past \cite{kafle2019challenges}. Such approaches employed the traditional early, late, and hybrid fusion \cite{baltruvsaitis2018multimodal}. They were unimodally pre-trained in the case of early and late fusion. However, some also multimodally pre-trained using hybrid fusion among \gls{cnn}-based visuals and an RNN or language model. Some have combined language models with visual information from images and videos at different levels of extracted language features starting at a word level, then to sentence level, and similarly from the paragraph to the end document level. The maximum amount of work was primarily focused on joining a word-level linguistic unit with features from images or videos for creating visual-semantic embeddings valuable on downstream applications. Additionally, numerous approaches are proposed built on n-grams, templates, and dependency parsing \cite{mogadala2019trends}. Furthermore, the encoder-decoder framework \cite{cho2014learning} became famous, which are image description based generation models. These are further extended with attention mechanisms \cite{bahdanau2014neural} to improve the harvesting of local image features, benefiting the word's initiation at each time step. Likewise, tasks such as \gls{vqa} \cite{antol2015vqa,goyal2017making} often consist of approaches that consist of an image feature extractor, a text encoder, a fusion module (normally with attention), and a response classifier.

Lately, the focus has been shifted to BERT based \gls{vl} models \cite{sun2019videobert} as BERT success has two keys: one is effective pre-training tasks over big language datasets, secondly, the use of Transformer \cite{vaswani2017attention} a contextualized text representations for learning instead of \gls{lstm}, which further pushed the \gls{vl} multimodal learning. Therefore, the focus has been shifted towards multimodal pre-training, which has brought leaping advances in \gls{vl} understanding tasks such as \gls{vqa} and \gls{vcr}, with great potential in extending to other \gls{vl} problems like memes classification, visual captioning, visual dialog, vision-language navigation, as well as video-and-language representation learning. Previously, most methods are designed for specific tasks, while BERT based \gls{vl} cross-modal such as VisualBERT \cite{li2019visualbert}, VilBERT \cite{lu2019vilbert}, LXMERT \cite{tan2019lxmert}, VL-BERT \cite{su2019vl}, B2T2 \cite{alberti2019fusion}, Unicoder-VL \cite{li2020unicoder}, ImageBert \cite{qi2020imagebert}, Pixel-BERT \cite{huang2020pixel} and UNITER \cite{chen2019uniter} has the ability to be fine-tuned to other downstream tasks. These BERT based \gls{vl} multimodal have attained state-of-the-art performance across diverse \gls{vl} problems, such as \gls{vqa}, \gls{vcr}, image-text retrieval and textual grounding.

Existing methods can be divided into two groups based on their model architecture. Fewer works like VilBERT \cite{lu2019vilbert} and LXMERT \cite{tan2019lxmert} utilize two-stream architecture based on the Transformer. The two-stream architectures process visual and language information respectively and fuse them afterward by another Transformer layer. On the other hand, there are methods such as B2T2 \cite{alberti2019fusion}, VisualBERT \cite{li2019visualbert}, Unicoder-VL \cite{li2020unicoder}, UNITER \cite{chen2019uniter}, Pixel-BERT \cite{huang2020pixel}, and VL-BERT \cite{su2019vl} which, apply single-stream architecture where two modalities are directly fused in the early stage and a single transformer is applied to both image and text modalities. They use BERT to learn a bi-directional joint distribution over the detection bounding box feature and sentence embedding feature. Further differences among them are in the training method, loss function, and datasets.

\subsubsection{Two-stream architecture}
VilBERT \cite{lu2019vilbert} and LXMERT \cite{tan2019lxmert} have both employed a two-stream architecture, and both visual and linguistic inputs are processed in separate streams. VilBERT proposed a co-attention mechanism that is also a transformer-based architecture. It allows vision attended language features to be integrated into visual representations and also vice versa by replacing key-value pairs in multi-head attention. It also permits flexible network depth for each modality and thus facilitates the cross-modal connections at various depths. VilBERT comprises of two parallel BERT-based models functioning over text segments and image regions. A third cross-modal where a succession of transformer blocks and co-attentional transformer layers is made to facilitate information exchange between modalities. It used Faster-R\gls{cnn} \cite{ren2015faster} and ResNet-101 \cite{he2016deep} as the backbone and are pre-trained on the Visual Genome dataset for regional feature extraction. Likewise, LXMERT proposed a similar kind of transformer model comprising three encoders: a language encoder, an object relationship encoder, and a cross-modality encoder. Its cross-modality encoder is slightly different from VilBERT as it comprises of two self-attention sub-layers, two feed-forward sub-layers, and one bi-directional cross-attention sublayer. It also utilized the Faster R-\gls{cnn} \cite{ren2015faster} for object detection. Other differences among these two approaches lie in the pre-training datasets, pre-training tasks, and downstream tasks after fine-tuning. A two-stream architecture's main issue is having a greater number of parameters with similar performance to a single stream architecture.

\subsubsection{Single-stream architecture}
Many other recent BERT-based approaches have opted for a single-stream architecture as it provides the same performance with fewer parameters \cite{chen2019uniter}. The single-stream model takes a mixed sequence of two modalities as an input. Many of the recent BERT based \gls{vl} models opted for single-stream architectures that include VisualBert \cite{li2019visualbert}, B2T2 \cite{alberti2019fusion}, Unicoder-VL \cite{li2020unicoder}, VL-BERT \cite{su2019vl}, Pixel-BERT \cite{huang2020pixel}, ImageBERT \cite{qi2020imagebert} etc. VisualBERT comprises several transformer layers piled such that it aligns features from an input text and regions that are extracted through Faster-RCNN, in the corresponding input image with a self-attention mechanism. They propose two VL model pre-training tasks such as sentence-image alignment and \gls{mlm}. They pre-trained their model on the coco caption dataset. They further evaluated their model on downstream tasks such as \gls{vqa}, \gls{vcr}, NLVR. B2T2 has proposed a similar architecture with the same pre-training tasks as \gls{mlm} and sentence-image alignment. However, it is evaluated on a single downstream task \gls{vcr}. The UNITER designs its model with four pre-training tasks, which are Image-Text matching, MLM, Masked region Modeling, and word-Region Alignment. They propose an image embedder and a text embedder to extract the corresponding embedding from image regions and tokens from a sentence. These embeddings are finally fed to the multi-layer transformer for cross-modality learning.

Freshly, many similar kind of cross-modal techniques has been proposed with a minor differences in architecture and fusion strategy, Pre-training tasks, Pre-training datasets, Pre-training strategy and downstream tasks such as ImageBERT \cite{qi2020imagebert}, Unicoder-VL \cite{li2020unicoder}, VL-bert \cite{su2019vl}, Pixel-BERT \cite{huang2020pixel}, InterBERT \cite{lin2020interbert}, B2T2 \cite{alberti2019fusion}, Vd-bert \cite{wang2020vd}, and many more. Some of them have used a similar model in different domains such as FashionBERT \cite{gao2020fashionbert} that deploy the text and image matching in cross-modal retrieval for the fashion industry. Some of them in video and text domains, like videoBERT \cite{sun2019videobert}, learn joint embeddings of video frame tokens and linguistic tokens from video-text pairs. CBT \cite{sun2019learning} presented contrastive learning to handle real-valued video frame features and others such as Univilm \cite{luo2020univilm} and ActBERT \cite{zhu2020actbert}. Villa \cite{gan2020large} has proposed Large-Scale Adversarial Training for Vision-and-Language Representation Learning. Pixel-BERT \cite{huang2020pixel} suggests aligning image pixels with text in contrast to conventional bottom-up features. HERO \cite{li2020hero} proposed hierarchical Transformer architectures to leverage both global and local temporal visual-textual alignments. VLP \cite{zhou2020unified} introduced pre-training tasks via the manipulation of attention masks for image captioning and \gls{vqa}. Multi-task learning was recently used in \cite{lu202012} to boost the performance further and enhance fine-tuning by using detected image tags.

\section{Research Issues, Opportunities, and Future Directions}
\label{sec:ResearchIssues}
%\subsection{Memes Classification}
\textbf{Memes Classification:} 
Generally, memes can be classified into two categories that are hateful/non-hateful. However, the issue arises in defining the hate in memes. Sometimes the margins are too narrow in memes identifying from hilarious to hateful. Usually, hate can be defined as an attack on people's characteristics like race, ethnicity, religion, sexual orientation, and many more \cite{kiela2020hateful}. Endorsing hateful memes can also be categorized as hateful. Furthermore, we can assume that memes can be further classified into many subcategories targeting the relevant issues. The recent black lives matter protest is a classic example of putting down a direct or indirect attack on someone’s race and color. Similarly, a new trend in social media in statuses where the certain text is written on colored background images can be tracked down into many categories like hateful/non-hateful, rumor, fake news, extremist, etc. Therefore, memes classification further into subcategories will also need to focus on research communities into sub-task-specific datasets and approaches such as fake, true or lies, propaganda memes, especially during an election. 

\textbf{Memes Reasoning:}
Further research issues will be to understand the semantic, especially those based both equally on text and images as into different categories, like whether the memes are humorous or hateful. Further memes categories, as described above, can also be elaborated in providing the possible relationship among different objects from the images and that of the text associated with it. It can also be seen as a problem like classification, and detection and segmentation of images can be aligned and form different elaborations of the images. However, identifying the most suitable alignment of detected objects and regions to that of the text part on that meme can be very challenging and may enhance the Vision and language multimodal ability to such a level that it can increase the model general ability further for other generalize Vision and language tasks.

\textbf{Memes Semantic Entailment:}
Another research issue can be memes semantic entailment. It is to predict if image semantically entails the text from the memes or independent of one another. This will be good in the sense like the example elaborated above where statuses on social media are frequently uploaded. Some user points of view or opinions have been shared on some color backgrounds. This will tell that if the visual and text are not entailing, an independent Text or image model from state-of-the-art can successfully classify the meme. Nevertheless, if semantic entails from image to the text have been found, it can be further processed by the multimodal like in the above two tasks for further categorization. 

\textbf{Multimodal fusion and co-learning:} Another vital research issue is related to the multimodal fusion and co-learning of visual and linguistic models. Traditionally, the researchers used many fusion strategies for a multimodal model wherein some have translated the modality into some uniformly feature set and then was fused into the \gls{ml} model in some cases, individually model has been trained. In the final stage, their decision has been fused \cite{baltruvsaitis2018multimodal}. However, recent advancements in Deep learning on Vision and NLP have brought the concept of co-learning to the next level by introducing the cross-modality training. Thus a hybrid fusion technique in the middle of the multimodal model has been prominent, and thus they are simultaneously trained on a cross-modal module. Thus a new concept of multimodally trained as unimodally and fused and multimodally trained as multimodally and a hybrid fusion has emerged along unimodally trained early and late fusion.

\section{Conclusion}
\label{sec:Conclusion}
On the rise of the Web, memes are regularly uploaded that need automatic censoring to hinder hate. Researchers from vision and language are inclining to \gls{vl} multimodal problems of which memes classification is picking the pace. Recent \gls{ml} methods performed significantly on \gls{vl} data but fail on memes classification. In this context, we presented an inclusive study on memes classification, generally on the \gls{vl} multimodal problems and current solutions. We further proposed a generalized framework for \gls{vl} problems. We also covered the early and next-generation works on \gls{vl} problems. Furthermore, we articulated several open research issues and challenges intending to guide the \gls{ml} research community for further investigation.

\section*{Acknowledgement}
{
This work was supported by the Institute for Information and Communications Technology Promotion Grant through the Korea Government (MSIT) under Grant R7120-17-1007 (SIAT CCTV Cloud Platform).
}


\printglossary
%\bibliographystyle{splncs04}
%\bibliography{references}

\end{document}